\pdfoutput=1
\documentclass[11pt]{article}

\usepackage{naacl2021}

\usepackage{times}
\usepackage{latexsym}

\usepackage[T1]{fontenc}
\usepackage[utf8]{inputenc}

\usepackage{microtype}

\usepackage{tikz}

\usepackage{xcolor}
\usepackage{amsmath}
\usepackage{booktabs} % For formal tables
\DeclareMathAlphabet\mathbfcal{OMS}{cmsy}{b}{n}

\usepackage[linesnumbered,ruled,vlined]{algorithm2e}

\SetCommentSty{mycommfont}
\DeclareMathAlphabet\mathbfcal{OMS}{cmsy}{b}{n}

\usepackage{lipsum}

\usepackage{multirow}
\usepackage{multicol}

\usepackage{breqn}

\usepackage{graphicx}

\usepackage{subcaption}

\usepackage{todonotes} % Insert [disable] to disable all notes:

\newcommand{\var}{\texttt}

\DeclareMathOperator*{\argmin}{arg\,min}
\usepackage{adjustbox}
\usepackage{graphicx}
\usepackage{wasysym}
\usepackage{tikzsymbols}

\title{Improving Autoregressive Training with Dynamic Oracles}

\author{
  Jianing Yang$^{\heartsuit\clubsuit}$\thanks{\ \*  Work done at CMU.} 
  \quad \quad \quad  \quad Harshine Visvanathan$^\clubsuit$ \\ \textbf{Yilin Wang}$^{\diamondsuit\clubsuit*}$ \quad \quad \quad \quad \textbf{Xinyi Hu}$^\clubsuit$ \quad \quad  \quad \quad \textbf{Matthew R. Gormley}$^\clubsuit$\\ \\
  $^\heartsuit$University of Michigan  \quad \quad   $^\clubsuit$Carnegie Mellon University \quad \quad    $^\diamondsuit$Harvard University\\
        {\tt \small jianingy@umich.edu}
}
\begin{document}
\maketitle

\begin{abstract}
Many tasks within NLP can be framed as sequential decision problems, ranging from sequence tagging to text generation. 
However, for many tasks, the standard training methods, including maximum likelihood (teacher forcing) and scheduled sampling, suffer from exposure bias \citep{ranzato2016sequence} and a mismatch between metrics employed during training and inference. DAgger \citep{ross2011reduction} provides a solution to mitigate these problems, yet it requires a metric-specific dynamic oracle algorithm, which does not exist for many common metrics like span-based F1, ROUGE, and BLEU. In this paper, we develop these novel dynamic oracles and show they maintain DAgger's no-regret guarantee for decomposable metrics like span-based F1. We evaluate the algorithm's performance on named entity recognition (NER), text summarization, and machine translation (MT). While DAgger with dynamic oracle yields less favorable results in our MT experiments, it outperforms the baseline techniques in NER and text summarization. 
\end{abstract}

\section{Introduction}

% Most of the key advances in NLP in recent years have come from a fundamental idea in transfer learning: pre-training an over-parameterized model on unlabeled text using a self-supervised objective and then fine-tuning on a downstream task \cite{clark_electra_2020, devlin_bert_2019, liu_roberta_2019, peters_deep_2018, raffel_exploring_2019, yang_xlnet:_2019}. The contextualized word embedding models have ranged in form from LSTM-LMs (e.g. ELMO \cite{peters_deep_2018}) to attention-based models (e.g. BERT \cite{devlin_bert_2019} and its descendants).

% \Matt{Starting with seq2seq feels a bit dated. That is, we should probably talk about learning for autoregressive models, which includes both seq2seq and decoder-only models. I think it's fine that our experiments focus on explicitly on seq2seq models. This might also suggest that we should change the title / abstract.
% Many tasks in NLP can be formulated as sequential decision problems.
% }

Numerous NLP tasks, ranging from sequence tagging to text generation, can be framed as sequential learning problems.  Frequently, algorithms tailored for such sequence learning employ an \emph{autoregressive} decoder. where the decoder recurrently uses its own output sequence as a prefix to decode the next token. In the conventional training paradigm known as teacher forcing, the decoder utilizes the \textit{ground truth} sequence as the left context for predicting the next token, as opposed to using its \textit{own} previous output as context---maximum likelihood (MLE) training is teacher forcing with cross-entropy loss. Teacher forcing faces two key challenges. 
The first is \textit{exposure bias}: during inference, the model is deprived of the ground truth for context and instead has to depend on its preceding predictions, which might be flawed. Because the decoder is not exposed to such flaws during training, it may struggle to rectify such errors during inference. 
Second, as highlighted in prior research \cite{RanzatoCAZ15, wiseman-rush-2016-sequence}, there's a discrepancy between the training loss, encourages the model to align outputs with the ground truth, and the inference metric, which could be based on a non-differentiable metric, like span-based F1, ROUGE, BLEU, among others.

% The seq2seq model \citep{sutskever_sequence_2014} has been widely applied to many NLP tasks that can be encoded as having sequences for both input and output, such as speech recognition, text summarization, image captioning, etc. Models developed under this framework typically involve an encoder and a decoder. The encoder captures relations among elements in the input sequence and encodes them into latent representations; the decoder consumes these latent representations and learns to produce sequences that are in the target sequence domain. However, despite the proliferation of architecture designs of seq2seq models, one of the key deficiencies of the existing seq2seq literature is the lack of good training algorithms. 

% Many tasks in NLP have benefited from advancements in architecture design for seq2seq models. These encoder-decoder models are often categorized by their underlying building blocks: LSTMs \cite{sutskever_sequence_2014}, CNNs \cite{gehring-etal-2017-convolutional}, or Transformer \cite{vaswani_attention_2017}. Many models widely used today have roots in the seq2seq design regime; combined with larger data and advancements in computation power, these models have been the driving force for the success of much NLP research. However, despite the proliferation of architecture designs of seq2seq models, one of the key deficiencies of the existing seq2seq literature is the lack of good training algorithms. Here, we give a preview of this challenge which motivates this paper.

\begin{figure}[]
    \subfloat[][]{%
    \includegraphics[width=0.8\linewidth]{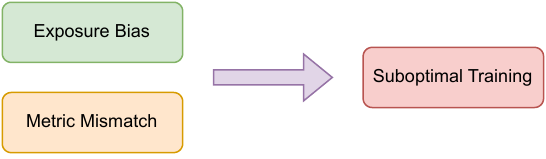}
}

\vspace{-4em}

\subfloat[][]{%
\resizebox{0.95\columnwidth}{!}{%
    \centering
\begin{tabular}{c  c  c  c  c}
    \bf Method & \bf Prefix & \bf Supervision & \adjustbox{angle=80,lap=\width-0.5em}{\bf Exposure Bias} & \adjustbox{angle=80,lap=\width-0.5em}{\bf Metric Mismatch} \\
    \toprule
    
    Teacher Forcing & Gold & Gold & \Sadey[1.4][red!60!white] & \Laughey[1.4][green!60!white] \\
    
    Scheduled Sampling & Gold + Predicted & Gold & \Laughey[1.4][green!60!white] & \Sadey[1.4][red!60!white] \\
    
    DAgger & Gold + Predicted & Dynamic Oracle & \Laughey[1.4][green!60!white] & \Laughey[1.4][green!60!white] \\

\end{tabular}

% \begin{tabular}{c  c  c  c}
%     \bf Method & \bf Prefix & \bf Supervision & \bf Problem \\
%     \toprule
    
%     Teacher Forcing & Gold & Gold & Exposure Bias \\
    
%     Scheduled Sampling & Gold + Predicted & Gold & Metric Mismatch \\
    
%     DAgger & Gold + Predicted & Dynamic Oracle & Doesn't exist yet \\

%     \end{tabular}
}
}
    
        \caption{\textbf{(a)} 
%         A dynamic oracle is a function that answers the question:
% Given a partial output sequence, what is the completion that minimizes loss with respect to the gold output?
Illustration of issues encountered in standard sequence training and the consequence. 
\textbf{(b)} Problems faced by Teacher Forcing, Scheduled Sampling and DAgger.}

    \label{fig:teaser}
    \vspace{-1.5em}
\end{figure}

\begin{figure*}[t]
    \centering
    \includegraphics[width=\linewidth]{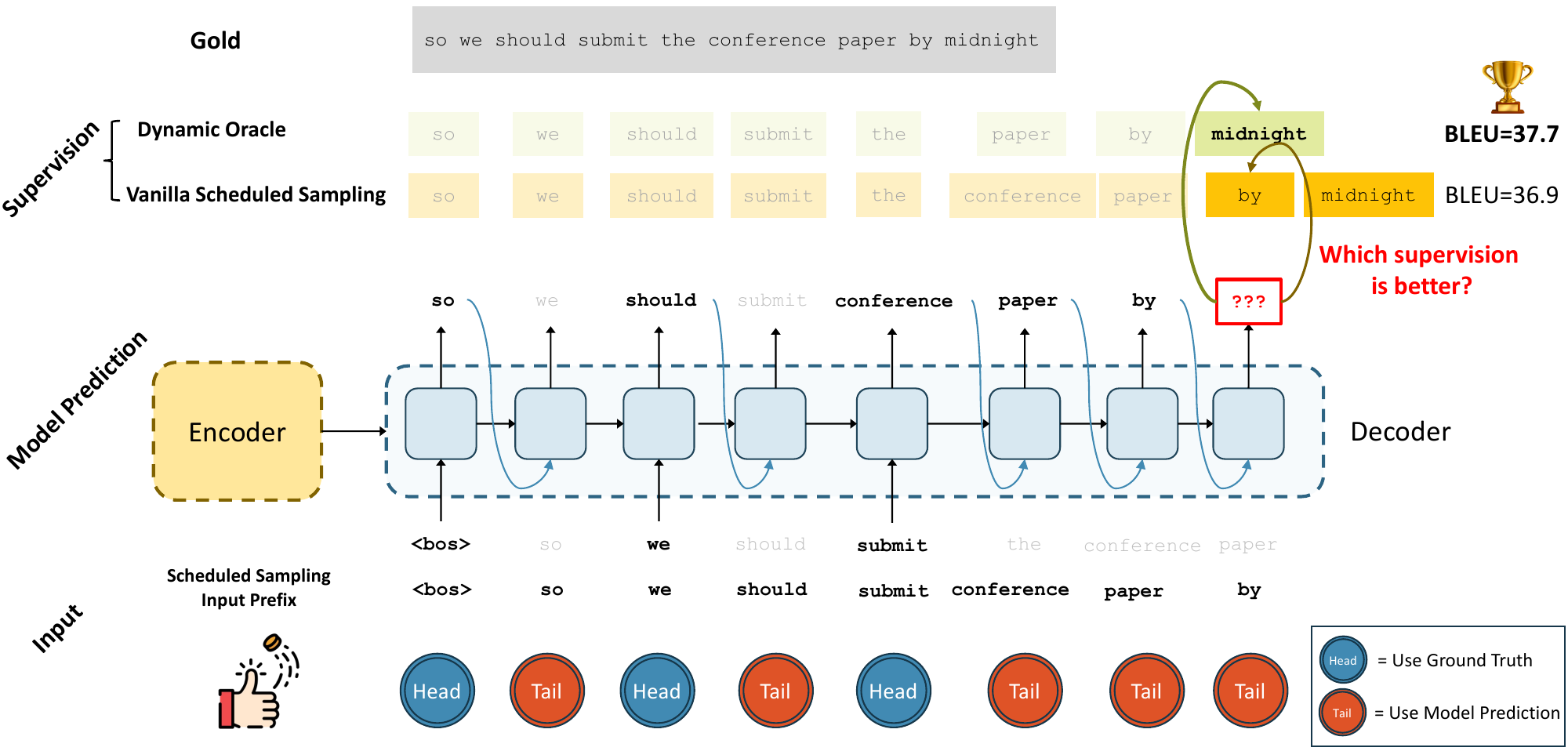}
    \caption{Dynamic oracle produces better supervision than vanilla scheduled sampling for training autoregressive decoder. In the above example, when using scheduled sampling to train a decoder, the word ``the" was not predicted by the decoder in the prefix sequence. In this case, what supervision should one use for the red box? Vanilla scheduled sampling uses the supervision ``so we should submit conference paper \textbf{by by midnight}" (notice ``by" appears twice) which leads to a BLEU of 36.9; whereas dynamic oracle uses the supervision ``so we should submit conference paper \textbf{by midnight}", which leads to a BLEU of 37.7. \textbf{The dynamic oracle gives better supervision.} Detailed explanation on scheduled sampling: at each autoregressive decoding step, one flips a coin to decide if a ground-truth token or a model-predicted token should be used as the prefix. As a result, the token ``the" happens to be not in the input prefix.}
    \label{fig:overview}
    \vspace{-1em}
\end{figure*}

% At test time, the decoder works in an \textit{autoregressive} manner, meaning the decoder recurrently feeds its own output sequence as a prefix to the decoder, to let it decode the next token.
% Note how the decoder depends \textit{solely} on itself at test time -- this is not true at train time. 
% In the conventional training paradigm, often denoted as Maximum Likelihood Estimation (MLE) or teacher forcing, the decoder utilizes the \textit{ground truth} sequence, as opposed to using its \textit{own} previous output as context, as the left context for generating the next token. 
% However, a significant divergence arises during inference, where the model must generate sequences without access to the ground truth reference context. The model must instead rely on its own preceding predictions, which might contain errors. As the decoder has not been explicitly trained to handle potentially flawed preceding contexts, it may struggle to rectify errors encountered during inference. 
% This disparity between training and testing scenarios, known as \textit{exposure bias}, highlights a substantial challenge inherent in sequence generation tasks.

Scheduled sampling \cite{bengio2015scheduled} aims to mitigate exposure bias by employing a prefix that is sampled from an interleaving of the ground truth and model-generated output. However, scheduled sampling still relies on the ground truth as output supervision, even when the input prefix is no longer the ground truth. This approach can lead to suboptimal supervision during training, as depicted in Figure \ref{fig:overview}. For instance, when the word ``the'' is omitted from the left context, using the gold sequence as supervision may not be the most appropriate choice. We argue that in cases where the left context is dynamically generated, it's imperative to employ a dynamically generated oracle for supervision rather than relying on the static ground truth sequence.

To tackle these challenges, we propose to use DAgger  \citep{ross2011reduction} in conjunction with novel dynamic oracles (cf. \S\ref{sec:dagger} and \S\ref{sec:dynamic-oracle}). These dynamic oracles are tailored to the specific test-time evaluation metric, and they serve as the optimal form of supervision, guiding the model in responding to errors made during decoding. However, for several loss functions, no known dynamic oracle exists, including examples such as span-based F1, ROUGE, and BLEU. In this study, we aim to bridge this gap. Our main contributions are:
\begin{itemize}
    \item We propose an algorithm to compute the \emph{exact} dynamic oracles for \emph{decomposable} metrics such as variants of span-based F1 (\S\ref{sec:f1-do}). DAgger with this exact dynamic oracle provides a no-regret guarantee. 
    \item For \emph{non-decomposable} metrics such as BLEU \citep{papineni-etal-2002-bleu} or ROUGE \citep{lin2004rouge}, we propose an algorithm to compute the \emph{approximate} dynamic oracle (\S\ref{sec:rouge-bleu-do}). 
    \item We evaluate the proposed algorithms on named entity recognition (NER) using the partial-match F1 metric (\S\ref{sec:ner-experiments}), machine translation (MT) with BLEU (\S\ref{sec:mt-experiments}), and summarization with ROUGE (\S\ref{sec:summarization-experiments}). Our MT experiments with BLEU indicate that DAgger may not consistently outperform robust baselines. However, our results on partial match F1 for NER and ROUGE for summarization, demonstrate the superiority of DAgger with our dynamic oracles, which surpass commonly used training methods like teacher forcing and scheduled sampling. 
\end{itemize}

\section{Methods}
\label{sec:methods}
We begin with an overview of the DAgger algorithm and the role of dynamic oracles before presenting our novel exact dynamic oracles for exact match F1/ partial match F1 and our new approximate dynamic oracles for ROUGE/BLEU.

\subsection{DAgger}
\label{sec:dagger}
%The DAgger algorithm \citep{ross2011reduction} iterates two procedures: (1) optimizes for a new policy to mimic a set of expert trajectories, and (2) uses the policy to collect additional trajectories, which are added to the existing expert trajectories. When DAgger is implemented with stochastic gradient descent (SGD), this aggregation process essentially equates to updating based on the most recently observed training example. In this context, DAgger, when combined with SGD and cross-entropy loss, defines a loss function that seeks to maximize the likelihood of dynamic oracle completions for each partial trajectory. This algorithm is designed to provide no-regret guarantees, ensuring it minimizes the knowledge of opportunities for superior actions that might have been missed. However, it necessitates a dynamic oracle customized to the specific task's loss function.

The DAgger algorithm \citep{ross2011reduction}, like other imitation learning algorithms, guides a model policy to be more like an expert policy. Applied to sequence generation, the model policy is simply the greedy decoder for an autoregressive model that selects the most proprobable next token $\hat{y}_t$ at each step to yield a sequence $\hat{\mathbf{y}}$. At each time step $t$, the expert policy (also called a dynamic oracle) examines both the ground truth sequence $\mathbf{y}^*$ and the current partial output of the model $\hat{\mathbf{y}}_{1:t-1}$ to return the completion of the output $\tilde{\mathbf{y}}_{t:T}$ that optimizes a task specific metric. 
During training, DAgger feeds in the model's own predictions $\hat{\mathbf{y}}$ to the decoder, which may have a different length than the ground truth. The DAgger loss function, if based on cross-entropy, then at each time step $t$ maximizes the likelihood of the first token $\tilde{y}_t$ from the expert policy's completion. When DAgger is implemented with an SGD-like optimization algorithm, we simply sum these likelihoods over all timesteps to get the loss for that example.
This differs from teacher forcing, which takes the ground truth $\mathbf{y}^*$ as both input to the decoder and as supervision. DAgger differs from scheduled sampling, which constructs a sequence whose length is the same as the ground truth by randomly interjecting some model prediction tokens $\hat{y}_t$ instead of the ground truth token $y^*_t$ and then uses the ground truth sequence $\mathbf{y}^*$ as supervision.
DAgger is designed to provide no-regret guarantees, ensuring it minimizes the knowledge of opportunities for superior actions that might have been missed. However, it requires a dynamic oracle customized to specific evaluation metrics

\subsection{Dynamic Oracles}
\label{sec:dynamic-oracle}

The term \emph{dynamic oracle}, originally introduced in the context of dependency parsing \cite{goldberg_training_2013},  represents the expert policy employed by DAgger. This dynamic oracle function acts as a guide to answer the question: \emph{Given a partial output sequence, which completion minimizes the loss when compared to the gold standard output?} It's important to note that dynamic oracles are inherently metric-dependent. In the context of sequence evaluation metrics, these can be broadly categorized into \emph{decomposable} and \emph{non-decomposable}. A decomposable metric can be expressed in the following form \cite{meister2020bestfirst}:
\begin{align}
    \text{score}(\mathbf{x}, \mathbf{y}) &= \sum_{t=1}^{N_{\mathbf{x}}} \text{score}( \mathbf{x}_{1:t-1} \circ x_t, \mathbf{y})
\end{align}\\
Here $\mathbf{y}$ is the ground truth sequence, $\mathbf{x}$ is the sequence to be evaluated and $N_{\mathbf{x}}$ is its length. 
%Metrics that cannot be written in this form are non-decomposable. In other words, 
Decomposable metrics, such as word error rate (WER) and span-based F1, can be broken down into additive components based solely on individual tokens (or $n$-grams) and the preceding partial sequence. Non-decomposable metrics cannot be broken down additively; e.g. BLEU or ROUGE both involve $n$-gram counts, but also global features (precision, brevity penalty, etc.) that prevent decomposition.

% We are interested in coming up dynamic oracles for both decomposable and non-decomposable metrics. We discover that it is usually easy to come up with \emph{exact} dynamic oracle algorithms for decomposable metrics.

For decomposable metrics, we define the \emph{exact} dynamic oracle to be the completion that gives the highest possible score (evaluated based on the metric) given a partial sequence. However, for non-decomposable metrics, such computation is usually computationally intractable; therefore, we propose to use an \emph{approximate} dynamic oracle, which is computed by approximation algorithm such as beam search. In the sequel, we give an exact dynamic oracle algorithm for span-based F1 (for chunking/NER) and an approximate dynamic oracle algorithm for ROUGE (for summarization) and BLEU (for machine translation).

\subsection{Exact and Partial F1}
\label{sec:f1-do}

\begin{algorithm}[t]
  \small
  \DontPrintSemicolon
  \SetNoFillComment
  \SetKwInOut{Input}{Input}
  \SetKwInOut{Output}{Output}
  \Input{$\mathbf{prev\_gold\_tag}$: previous gold tag,\\
   $\mathbf{curr\_gold\_tag}$: current gold tag \\
   $\mathbf{prev\_tag}$: previous predicted single tag}
  \Output{next best token\\}

%   \SetKwFunction{FMain}{DynamicOracleF1}
%   \SetKwProg{Fn}{Function}{:}{}
%   \Fn{\FMain{$\mathbf{prev\_gold\_tag}$, $\mathbf{curr\_gold\_tag}$, $\mathbf{prev\_tag}$}}{
    \uIf{start(\textup{curr\_gold\_tag}) = `B'}{
        \uIf{start(\textup{prev\_tag}) = `B' and type(\textup{curr\_gold\_tag}) = type(\textup{prev\_tag})}{
            \Return `I-' + type(\textup{curr\_gold\_tag})
        }
        \Else{
            \Return $\mathbf{curr\_gold\_tag}$
        }
    }
    \uElseIf{start(\textup{curr\_gold\_tag}) = `I'}{
        \uIf{\textup{prev\_tag} is `O'}{
            \Return `B-' + type(\textup{curr\_gold\_tag})
        }
        \uElseIf{type(\textup{prev\_tag}) $\neq$ type(\textup{prev\_gold\_tag})}{
            \Return `O'
        }
        \uElse{
            \Return $\mathbf{curr\_gold\_tag}$
        }
    }
    \uElse{
        \Return `O'
    }
%   }
\textbf{Note:} 
\begin{itemize}
    \small \item \textit{start}(tag) returns the prefix of the tag, i.e, \texttt{B}, \texttt{I}, or \texttt{O}.
    \small \item \textit{type}(tag) returns the entity type, i.e, \texttt{PER}, \texttt{LOC}, etc.
\end{itemize}
\caption{Dynamic Oracle for Partial F1 Score}
\label{algo:do_f1}
\end{algorithm}

Early work in chunking (CoNLL-2000 shared task \cite{tjong_kim_sang_introduction_2000}) and NER (CoNLL-2003 shared task \cite{tjong_kim_sang_introduction_2003})  established \emph{exact-match F1} as the primary metric, computing the harmonic mean of precision and recall based on exact spans in reference and system output. For a single sentence, this metric can be considered as a decomposable metric, with an exact dynamic oracle algorithm. We introduce a dynamic oracle algorithm for the exact F1 score given in Appendix Algorithm \ref{algo:do_f1_exact}.

% It's a decomposable metric, with an exact dynamic oracle algorithm in the Appendix. 
%However, exact match F1 is that it does not account for type mismatch or partial matches, 

% Early work in chunking (CoNLL-2000 shared task \cite{tjong_kim_sang_introduction_2000}) and NER (CoNLL-2003 shared task \cite{tjong_kim_sang_introduction_2003}) established F1 as the primary metric for evaluating span-based sequence tagging tasks. Here we refer to this metric as \emph{exact-match} F1 because it computes the harmonic mean of precision and recall based on only spans that exactly match (both in the span of tokens and the label) between the reference and system output. Because the F1 score is computed by the individual tags, it is clear that exact-match F1 is a decomposable metric. An exact dynamic oracle algorithm for exact F1 is given in the Appendix.

% A shortcoming of the exact match F1 is that it does not account for type mismatch or partial matches.
% The \texttt{SemEval’13 - 9.1 task} \cite{segura-bedmar-etal-2013-semeval}, introduces four different ways to measure the precision, accuracy, and the F1-score based on metrics defined in the Message Understanding Conference (MUC) \cite{chinchor-sundheim-1993-muc}. 
%
% This includes exact boundary and type matching; exact boundary matching regardless of type; partial boundary matching regardless of type; and, some overlap between the tagged entity and gold annotation is present. This opens a new way of evaluating the model prediction instead of just looking for an exact match. 
% % \matt{Add citations for SemEval'13 and MUC.}

A limitation of the exact match F1 metric lies in its failure to consider type mismatch and partial matches. Moreover, prior research \cite{manning_2006} suggests that optimizing for exact-match F1 scores may not be suitable for NER tasks, as it can lead to a system's reluctance to recognize entities, resulting in a prevalence of 'O' tags. Consequently, we develop dynamic oracles tailored for the \emph{partial} F1 metric, as defined in MUC \cite{chinchor-sundheim-1993-muc}. Algorithm \ref{algo:do_f1} presents our dynamic oracle for partial-match F1. Like the exact-match F1, partial F1 is decomposable, making its dynamic oracle algorithm exact. A formal proof of the algorithm's correctness is available in Appendix \ref{app:partial_f1_proof}.

This algorithm is designed to predict the next optimal token in NER tasks. Its simplicity derives from the nature of the count/divide approach utilized by the metric. This algorithm incorporates various conditions that account for each conceivable output. We use an example to illustrate the rationale behind this algorithm. If the tagging has already begun, with the same type, i.e., 
\begin{align*}
    \texttt{prev\_gold\_tag} &= \texttt{`O'} \\
    \texttt{curr\_gold\_tag} &= \texttt{`B-LOC'} \\ 
    \texttt{prev\_tag} &= \texttt{`B-LOC'}
\end{align*}
Teacher forcing suggests the optimal subsequent tag would be \texttt{B-LOC}, which would result in a false positive error. The Dynamic Oracle suggests \texttt{I-LOC}, which is the ideal succeeding completion to attain a favorable partial match F1 score.

The application of exact dynamic oracles isn't exclusive to the NER task or the partial F1 metric. Another metric that can benefit from such an approach is the word error rate (WER), which is prevalently used in automatic speech recognition (ASR) tasks. For completeness, we present the dynamic oracle algorithm and proof of correctness in Appendix \ref{app:wer}.

% , the ideal succeeding completion would be \texttt{I-LOC}, a prediction correctly advocated by the Dynamic Oracle. However, if we opt for the Teacher Forcing prediction instead, it necessitates the inclusion of a False Positive value in the F1 Score computation.

\subsection{ROUGE and BLEU}
\label{sec:rouge-bleu-do}

\begin{algorithm}[t]
  \small
  \DontPrintSemicolon
  \SetNoFillComment
  \SetKwInOut{Input}{Input}
  \SetKwInOut{Output}{Output}
  \Input{$\mathbf{x}_{1:i-1}$: input prefix sequence up to $i-1$,\\
   $\mathbf{y}$: gold sequence,\\
   score(seq1, seq2): scoring function
%   score: a sentence scoring function,\\
%   beam_size: the beam size,\\
%   beam_length: max length to expand beam,
   }
  \Output{$y^{D.O.}_i$: next word dynamic oracle}

    Initialize \texttt{candidates} to be a set of unigrams in $\mathbf{y}$\\
    
    Initialize \texttt{queue} to be a priority queue\\
    
    Insert $\mathbf{x}_{1:i-1}$ with score=$-\infty$ into \texttt{queue} \\

    \For(){$l = 1...\texttt{beam_length}$}{
        \For(){\texttt{prefix} $\in$ \texttt{queue}}{
            \For(){\texttt{word} $\in$ \texttt{candidates}}{

                \texttt{new_seq} $\gets$ \texttt{prefix} concatenated by \texttt{word}\\
                $s \gets$ score(\texttt{new_seq}, $\mathbf{y}$) \\
                Insert \texttt{new_seq} with score=$s$ into \texttt{queue} \\
            }
            Truncate \texttt{queue} by only keeping the top-\texttt{beam_size} scoring sequences
        }
    }
    $y^{D.O.}_i \gets$ select the top-1 scoring sequence from \texttt{queue} and get the $i$-th word \\
    \Return $y^{D.O.}_i$

 \caption{Approximate Dynamic Oracle for ROUGE/BLEU with Beam Search}
 \label{algo:rouge}
\end{algorithm}

The ROUGE score \citep{lin2004rouge} is a widely adopted metric for evaluating summarization tasks. ROUGE measures both $n$-gram precision and recall, culminating in the harmonic mean of these two scores. However, it is not decomposable due to the global nature of $n$-gram counts used in precision and recall calculations. Identifying the exact completion that maximizes ROUGE for a partial sequence is computationally prohibitive. Consequently, we employ beam search as an approximation method to search for the dynamic oracle in the context of ROUGE. In Algorithm \ref{algo:rouge}, we present an approximate dynamic oracle algorithm for ROUGE, where the optimal completion is determined by computing the ROUGE score for \texttt{beam_size} potential completions, each up to a length of \texttt{beam_length}, and selecting the completion with the highest score.

The BLEU score \citep{papineni-etal-2002-bleu}, is a widely-used metric in machine translation tasks. Like ROUGE, BLEU relies on global $n$-gram counts, making it a non-decomposable metric. Thus, we employ the same approximate dynamic oracle algorithm outlined in Algorithm \ref{algo:rouge} that utilizes BLEU as the scoring function. In our study, we consider BLEU-4, encompassing up to $4$-grams.

\section{Experiments}
To gauge the effectiveness of our approach, we conducted experiments on three sequence-to-sequence (seq2seq) modeling problems, including named entity recognition (NER), machine translation (MT), and summarization. 

\subsection{F1 / Named Entity Recognition}
\label{sec:ner-experiments}

\paragraph{Dataset} We used three NER benchmarks.
%: the CoNLL-2003 shared task and the WNUT 2017 shared task.
The CoNLL-2003 Shared Task \citep{tjong-kim-sang-de-meulder-2003-introduction} comprises formal writings sourced from the Reuters news article corpus; we evaluate on English and German.
The WNUT 2017 Shared Task \citep{derczynski2017results} focuses on extracting emerging named entities from noisy user-generated tweets.

\paragraph{Experiment Details}
\begin{table}[t]
\centering
\resizebox{\columnwidth}{!}{%

    \begin{tabular}{c c c}

        \bf Dataset+Model & \bf Training Method & $\mathbf{F1_p}$ (\%)\\
        \hline
        
        \multirow{3}{0.2\columnwidth}{\small CoNLL-03(De) + BERT} & Teacher Forcing  & 86.55 \\

          & Scheduled Sampling  & 86.72 \\

          & DAgger  & \bf 87.02 \\
        \midrule
        
        \multirow{3}{0.2\columnwidth}{\small CoNLL-03(En) + BERT} & Teacher Forcing  & 89.89 \\
          & Scheduled Sampling  & 89.98 \\
          & DAgger & \bf 90.21 \\
        \midrule
        \multirow{3}{0.2\columnwidth}{\small CoNLL-03(En) + FLAIR}  & Teacher Forcing  & 91.50 \\
         & Scheduled Sampling  & 92.72 \\
         & DAgger & \bf93.08 \\
        \midrule
        \multirow{3}{0.2\columnwidth}{\small WNUT-17 + FLAIR} & Teacher Forcing & 47.52 \\
          & Scheduled Sampling & 48.91 \\
          & DAgger  & \bf49.40 \\
        \hline
    \end{tabular}
}
    
    \caption{Partial match F1 ($\mathbf{F1_p}$) results on NER datasets. DAgger with exact dynamic oracle consistently outperforms teacher forcing and scheduled sampling across datasets and backbone models.}
    \label{tab:NER_Results}
    \vspace{-1.5em}
\end{table}

We follow prior work \cite{shen2017deep} in formulating the NER task as a greedy decoding problem.\footnote{This is in contrast to many prior approaches which use a linear-chain CRF layer and find the most probable sequence through dynamic programming.}
We evaluated our proposed learning method on two models:
\begin{description}
\item[BERT + RNN-LM] Our first model uses BERT-base \cite{devlin_bert_2019} encoder followed by an RNN-LM decoder \cite{shen2017deep}. 
\item[FLAIR + RNN-LM] Our second model uses pooled FLAIR embeddings \cite{akbik-etal-2019-pooled} fine-tuned in the encoder, followed by an RNN-LM decoder \cite{shen2017deep}. 
\end{description}
%
% It's worth noting that the RNN-LM decoder, as described by \citet{shen2017deep}, was applied to all of the above models.
% We compare three approaches to training the seq2seq model: Teacher Forcing, Scheduled Sampling, and DAgger with exact dynamic oracle for partial F1. 
% We experimented with batch sizes of 8, 16, and 32 and learning rates of 0.1, 0.5, and 1.0 and found that a batch size of 16 with a learning rate starting from 1.0 is optimal for the CoNLL-2003 dataset and a batch size of 32 with a learning rate starting from 0.5 is optimal for the WNUT-2017 dataset. 
%
We conducted a comparison of three training approaches for these seq2seq models, which included Teacher Forcing, Scheduled Sampling, and DAgger with an exact dynamic oracle for partial-match F1. We chose an initial learning rate of 1.0 and batch size of 16 for the CoNLL-2003 dataset and an initial learning rate of 0.5 and batch size of 32 for the WNUT-2017 dataset. These choices are based on hyper-parameter search, with potential batch size (8, 16, 32) and learning rate (0.1, 0.5, 1.0). 

% We experimented with different batch sizes (8, 16, and 32) and learning rates (0.1, 0.5, and 1.0). The results suggest that a batch size of 16, with an initial learning rate of 1.0, performed optimally for the CoNLL-2003 dataset. A batch size of 32 with an initial learning rate of 0.5 appears optimal for the WNUT-2017 dataset.

% The schedule for the scheduled sampling starts at the third epoch, after which 20\% of the predicted output is fed back into the model every epoch. 

For scheduled sampling and DAgger, we warm-start the model with several epochs of teacher forcing, then switch to scheduled sampling on the third epoch, with 20\% of the predicted output being fed back into the model every subsequent epoch. For the FLAIR experiments, we retrieve the embeddings, which include GLoVe Word Embeddings \citep{pennington-etal-2014-glove}, Character Embeddings, Forward Pooled FLAIR embeddings, and Backward Pooled FLAIR embeddings \citep{akbik-etal-2018-contextual}, from the FLAIR library \citep{akbik-etal-2019-flair}. The RNN-LM decoder consists of a single-layer LSTM with 50 hidden units. 

% % The schedule for the scheduled sampling starts at the third epoch, after which 20\% of the predicted output is fed back into the model every epoch. 
% The embeddings utilized were sourced from the FLAIR library \citep{akbik-etal-2019-flair}. These included GLoVe Word Embeddings \citep{pennington-etal-2014-glove}, Character Embeddings, Forward Pooled FLAIR embeddings, and Backward Pooled FLAIR embeddings \citep{akbik-etal-2018-contextual}.

% The RNN-LM decoder consists of a single-layer LSTM with 50 hidden units. 

\paragraph{Results}
The results are in Table \ref{tab:NER_Results}. For WNUT-17 with the FLAIR mdoel, scheduled sampling leads to a marked improvement over teacher forcing (+1.39 $\text{F1}_{\text{p}}$). However, the seq2seq model with the DAgger (dynamic oracle) training achieves a further improvement over scheduled sampling (+0.49 $\text{F1}_{\text{p}}$) and a correspondingly greater gap over teacher forcing (+1.88 $\text{F1}_{\text{p}}$). Similar, albeit more modest, trends are observed on the CoNLL-2003 benchmark with both the FLAIR and BERT models on both English and German. For example, DAgger (dynamic oracle) training with the FLAIR model outperforms both scheduled sampling (+0.36 $\text{F1}_{\text{p}}$) and teacher forcing (+1.58 $\text{F1}_{\text{p}}$). 

\subsection{BLEU / Machine Translation}
\label{sec:mt-experiments}

\paragraph{Dataset}
The IWSLT'14 \footnote{https://workshop2014.iwslt.org/} Sl-En dataset contains $17,815$ parallel Slovenian-English sentences. 
For our machine translation experiments, we opted for this relatively small dataset to enable the execution of a greater number of experiments and facilitate more comprehensive analysis. 

\paragraph{Experiment Details}

In addition to teacher forcing and scheduled sampling, we consider the following baseline methods: 

\begin{description}
\item[Minimum Risk] \citet{shen-etal-2016-minimum}  trains the model to minimize empirical ``risk'', defined as the weighted sum of the BLEU score for a batch of potential translations. Each score is weighted by the probability of the corresponding sample as predicted by the model.
\item[Word-Level Oracle] Proposed by \citet{zhang-etal-2019-bridging}, this method involves sampling context words from both the ground truth sequence and the sequence predicted by the model during training. It also added Gumbel noise \cite{gumbel1954statistical} to the models' prediction to enhance robustness, which diversifies prefixes feed into the model.
% he predicted sequence is selected based on sentence-level optimization.
\item[MIXER] \citet{ranzato2016sequence} employs any reward function to directly optimize the evaluation metrics used at test time.    
\end{description}
% We use a standard Transformer \cite{vaswani_attention_2017} encoder-decoder model and train from scratch %using fairseq (\citet{ott-etal-2019-fairseq}).
For each method, we used a standard encoder-decoder Transformer model \cite{vaswani_attention_2017} and trained it from scratch. % using the fairseq toolkit \cite{ott-etal-2019-fairseq}.

% For this task, when choosing the sampling schedule for scheduled sampling and DAgger, we wanted to explore the effect of starting sampling in the late stage of training. As suggested in \citet{RanzatoCAZ15}, starting diversifying the prefix from a well-trained stage might help the model find the optimal policy quicker. Therefore, for this task, we first train the model using teacher forcing for 167 epochs, then start sampling at epoch 168 for scheduled sampling and DAgger. Epoch 167 was chosen because it gave the best BLEU ($11.67$) on the validation set among the first 300 epochs of teacher-forcing training. We then train for 5 epochs for each setting listed in Table \ref{tab:BLEU_results}. 

In this task, we aim to investigate the impact of initiating sampling during the later stages of training. As proposed by \citet{RanzatoCAZ15}, introducing diversity into the prefix during the later training phases may expedite the model's search for the optimal policy. In this experiment, we first train the model using teacher forcing for 167 epochs. We then train the model using each aforementioned technique for 5 epochs. We chose epoch 167 because it produced the highest BLEU score ($11.67$) on the validation set among the initial 300 epochs of teacher-forcing training. 
% Following this, we conducted training for a duration of 5 epochs for each configuration detailed in Table \ref{tab:BLEU_results}.

% However, results on this setting suggest that starting at such a late epoch does not work well for our DAgger method. We subsequently conduct analysis to root-cause the negative results and describe our findings in Section \ref{sec:mt-results}

\paragraph{Results}
\label{sec:mt-results}

\begin{table}[t]
    \centering
\resizebox{0.75\columnwidth}{!}{%
    \begin{tabular}{c c}

        \bf Training Method & \bf BLEU \\
        \hline
        Epoch 167 (baseline) & 11.67 \\
        \hline
        Teacher Forcing & 11.28 \\
        Minimum Risk & 11.21 \\
        Word-Level Oracle & \textbf{12.08} \\
        MIXER & 12.04\\ 
        Scheduled Sampling  & 11.25  \\\hline 
        DAgger + Greedy Search & 11.44 \\
        DAgger + Beam Search, size=3 & 11.16 \\
        DAgger + Beam Search, size=5 & 11.24 \\
        \hline
    \end{tabular}
}
    \caption{BLEU results on IWSLT '14 Sl-En. Size indicates beam\_size used in Algorithm \ref{algo:rouge}. All methods except the word-level oracle and MIXER fail to outperform the baseline. }
    \label{tab:BLEU_results}
    \vspace{-1.5em}
\end{table}
\begin{figure}[t]
    \centering
    \subfloat[][]{%
    \includegraphics[width=0.95\linewidth]{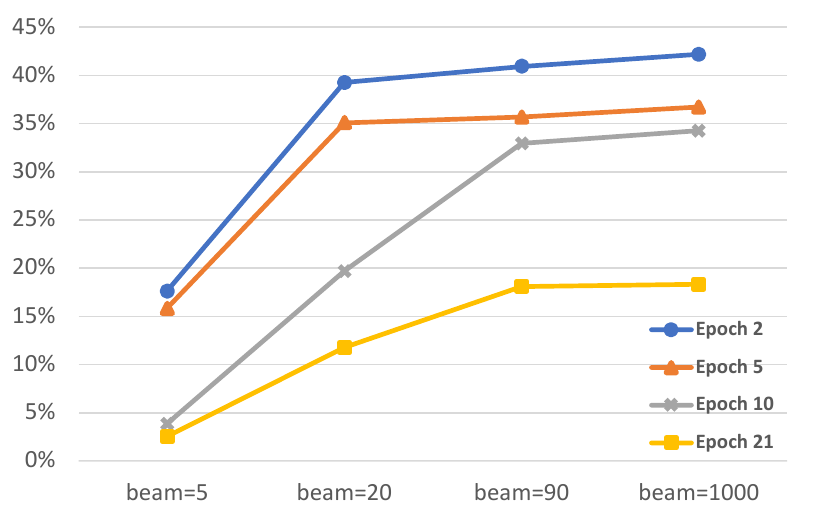}
    }

    % \vfill
    % \vspace{-1em}

    \subfloat[][]{%
        \resizebox{\columnwidth}{!}{%
    \begin{tabular}{c | c | c c c c}

        \multirow{2}{*}{\bf Epoch} & \multirow{1}{*}{\bf Ground Truth} &  \multicolumn{4}{c}{\bf Dynamic Oracle ($\Delta$ BLEU)}\\
         & \bf (BLEU) & \bf 5 & \bf 20 & \bf 90 & \bf 1000\\
        \hline
        2 & 72.10 &  +2.98 & +4.84 & +4.97 & +5.04 \\
        5 & 74.66 & +2.33 & +4.19 & +4.3 & +4.37 \\
        10 & 78.52 & +0.24 & +0.89 & +1.23 & +1.31 \\
        21 & 86.42 & +0.11 & +0.38 & +0.52 & +0.55 \\
    \end{tabular}
    }
    }
    
    \caption{Larger beam sizes and initiating DAgger training earlier results in better dynamic oracle quality. \textbf{(a)} The vertical axis indicates the percentage of instances where the dynamic oracle's completion yields a \textit{higher} BLEU score than that of the ground truth. A higher percentage implies greater benefits from transitioning from teacher forcing to the dynamic oracle. Importantly, by design, the dynamic oracle's BLEU cannot be lower than the ground truth--it always selects the better word between the ground truth and the beam search result. \textbf{(b)} Comparison of average BLEU scores for ground truth and dynamic oracle supervisions, averaged over an entire batch. Beam sizes of 5, 20, 90, and 1000 are employed for the dynamic oracle.
}
\vspace{-2.5em}
    \label{fig:bleu_analysis}
\end{figure}

% However, the outcomes under this setup indicated that starting at such a late epoch was less effective for our DAgger method.

As shown in table \ref{tab:BLEU_results}, DAgger with dynamic oracle seems less favorable compared to the baseline and other methods. It is noteworthy that other methods also have no or very marginal improvement compared to the baseline. It is possible that BLEU has saturated at this point. 

\paragraph{Analysis}
We present further analysis in Figure \ref{fig:bleu_analysis} to study if this phenomenon is related to our decision to initiate DAgger from a relatively late epoch. Specifically, we seek to answer: given a prefix, how does the BLEU score of a sentence, completed by the dynamic oracle, compare with one that's completed by copying the ground truth starting from the same prefix position? We consider two key metrics: (1) \emph{How many} times does dynamic oracle give higher BLEU than ground truth? (2) \emph{how much} better, in terms of average BLEU, does dynamic oracle achieve, compared to ground truth?

This analysis offers two insights: (1) Starting DAgger in \emph{early} epochs gives better dynamic oracles (in terms of the above two metrics). (2) using \emph{larger} beam sizes gives better dynamic oracles. However, it's worth noting that this improvement begins to plateau beyond a beam size of 20.

These insights partially explain the unsatisfactory performance of DAgger when starting from epoch 168 with a beam size of 3 or 5.  Insight (1) is evidenced by comparing the two metrics while varying the starting epoch but keeping the beam size fixed—both the frequency of BLEU improvement and the magnitude of BLEU improvement remain relatively low when starting from a later epoch. Insight (2) is evidenced by keeping the starting epoch constant and varying the beam size, where the general trend reveals that larger beam sizes are more beneficial, though the improvements begin to diminish beyond a beam size of 20. It may be that carefully tuning the use of larger beam sizes and starting DAgger training in earlier epochs could yield better results.

% Consequently, we carried out an in-depth analysis to identify the root causes behind the unfavorable results, and these findings are discussed in Section \ref{sec:mt-results}.

% Although we hypothesized initializing with a well-trained model could be beneficial to DAgger training, as shown in Table \ref{tab:BLEU_results}, DAgger with dynamic oracle, as well as other training methods, does not improve BLEU with this initialization. 
% It appears that BLEU has saturated at this point. 
% Therefore 

\subsection{ROUGE / Summarization}
\label{sec:summarization-experiments}

\paragraph{Dataset}
For summarization, we use the CNN/Daily-Mail (CNNDM) \citep{hermann2015teaching,nallapati2016abstractive} benchmark, which contains news articles and associated highlights as summaries. 
%Summaries here are typically closely related to source sentences.

\paragraph{Experiment Details}

\begin{table}[t]
\centering
\resizebox{\columnwidth}{!}{%

    \begin{tabular}{c c c c}

        \bf Training Method & \bf ROUGE-1 & \bf ROUGE-2 & \bf ROUGE-L \\
        \hline
        Teacher Forcing & 41.30 & 19.59 & 37.86\\
        Scheduled Sampling & 42.11 & 20.02 & 38.21 \\
        DAgger + Greedy Search & 42.35 & 20.49 & 38.3 \\
        DAgger + Beam Search 2 & 42.55 & 20.72 & 38.53 \\
        DAgger + Beam Search 3 & 42.49 & 20.65 & 38.59 \\
        DAgger + Beam Search 5 & \bf 42.62 & \bf 20.79 &  \bf 38.64 \\
        \hline
    \end{tabular}
}
    \caption{DAgger with dynamic oracle training improves ROUGE results on CNN/DailyMail summarization task. The dynamic oracles optimize the ROUGE-2 metric. The number following ``beam search" is the beam size. DAgger with beam search size 5 consistently outperforms teacher forcing and scheduled sampling across all ROUGE metrics. We ran a paired permutation test on the results between DAgger and scheduled sampling on the ROUGE-2 metric. The p-value is 0.055, which suggests statistical significance. 
    }
    \label{tab:ROUGE_results}
    \vspace{-1.5em}
\end{table}

The baseline system for our experiments is BART \cite{lewis2019bart}, an encoder-decoder model consisting of a BERT-style encoder followed by a pre-trained decoder (GPT-2). We use fairseq's BART \citep{wang2019espresso}.

% We compare three approaches to train the seq2seq model: Teacher Forcing, Scheduled Sampling, and DAgger with approximate dynamic oracle for ROUGE.

We compare three methods for training BART: Teacher Forcing, Scheduled Sampling, and DAgger with our approximate dynamic oracle for ROUGE. We experimented using two configurations of the approximate dynamic oracle algorithm. The first, referred to as "greedy search," sets the \texttt{beam_length} parameter to 1. The second, known as ``beam search," sets the \texttt{beam_length} to 2, 3, and 5. 

% We experimented with two different settings of approximate dynamic oracle algorithm, one we call \textit{greedy search}, where we set \texttt{beam_length} to be $1$  and one we call \textit{beam search}, where we set the \texttt{beam_length} to be $5$.  For \textit{beam search} we also try with beam sizes 2, 3, and 5.

% The sampling schedule for scheduled sampling starts at epoch 2, after which the decoder's output is fed 25\% of the time back to the decoder as a prefix. An identical sampling schedule (starting from epoch 2, sample at 25\%) is used for DAgger. The model is run for 15 epochs for every training setup. 

For scheduled sampling and DAgger, we start the algorithm at epoch 2. During this process, the decoder's output is incorporated back into the decoder as a prefix 25\% of the time. We run all experiments for 15 epochs.

\paragraph{Results}
From the results in the table \ref{tab:ROUGE_results}, it's evident that the model trained with DAgger (Dynamic Oracle) surpasses those trained with teacher forcing and scheduled sampling for both the greedy search and beam search approximated oracles. The best performance is observed when employing a beam search approximation with a beam size of 5 in conjunction with the DAgger algorithm, resulting improvements across ROUGE-1, ROUGE-2, and ROUGE-L scores over both scheduled sampling (+0.51 R-1, +0.77 R-2, +0.43 R-L) and teacher forcing (+1.32 R-1, +1.20 R-2, +0.78 R-L).

\section{Related Work}
\label{background}

\subsection{Dynamic Oracle for Other Metrics}
Dependency parsing has a rich history of dynamic oracle training with algorithms that predate are analogous to DAgger \cite{ross2011reduction}.  Transition-based, deterministic, tabular parser with non-deterministic dynamic oracle has been studied on the labeled attachment score (LAS) and the unlabeled attachment score (UAS) \cite{gomez-rodriguez_polynomial-time_2014,gomez-rodriguez_efficient_nodate}. Dynamic oracles are also used for projective and non-projective parsing \cite{gomez-rodriguez_polynomial-time_2014,gomez-rodriguez_efficient_nodate}. Apart from dependency parsing, dynamic oracles are also designed for Word Error Rate (WER) and constituency parsing \cite{sabour2018optimal,Fried2018PolicyGA}. Yet, this line of work does not offer a solution for non-decomposable metrics.
%
% \citet{goldberg_training_2013, goldberg_dynamic_2012,goldberg_tabular_2014} utilized transition-based, deterministic, tabular parser with non-deterministic dynamic oracle on the labeled attachment score (LAS) and the unlabeled attachment score (UAS), \citet{gomez-rodriguez_polynomial-time_2014,gomez-rodriguez_efficient_nodate} focused on dynamic oracles projective and non-projective parsing. 
% \citet{goldberg_dynamic_2012} introduced dynamic oracles on deterministic dependency parsing for the arc-eager transition system. \citet{goldberg_training_2013,} brought training deterministic dynamic oracles for non-deterministic metrics in dependency parsing.

% \Matt{Add citations for dynamic oracles for other metrics: dependency parsing. See slack}  
These dynamic oracles can also be used in a more recent variant of DAgger that takes the beam-search approximation into account \cite{negrinho_learning_2018, negrinho_empirical_2020} and takes interactive no-regret into account \cite{ross2014reinforcement}.  

% \Matt{Add citations on Renato's work here. Other possible variants of DAgger include Aggrevate (Ross et al., 2014)}

% For some task-specific losses and hyperparameter choices, scheduled sampling behaves the same way as DAgger. Notably, when applied to seq2seq models, scheduled sampling can be seen as DAgger training using a hamming loss rather than the actual task-specific loss (e.g., F1, BLEU, ROUGE, WER). This means that scheduled sampling is optimizing for the "wrong" loss function.

% In our research, we employ DAgger training to address this limitation in the standard practice of training seq2seq models. To do this, we need to create dynamic oracles for the loss functions we are interested in.

\subsection{Methods Based on Scheduled Sampling}

Scheduled sampling has been improved in various ways. There exist differentiable versions of scheduled sampling \cite{goyal_differentiable_2017, xu_differentiable_2019} and faster versions achieved via parallelization \cite{duckworth_parallel_2019, mihaylova_scheduled_2019}. It could also be extended to allow more diverse prefixes by adding noise to models' predictions \cite{zhang-etal-2019-bridging}

% \citet{goyal_differentiable_2017, xu_differentiable_2019} proposed methods to make scheduled sampling differentiable. 
% \citet{duckworth_parallel_2019, mihaylova_scheduled_2019} suggested methods to accelerate scheduled sampling through parallelization. 
% \citet{zhang-etal-2019-bridging} extended scheduled sampling by allowing more diverse prefixes to be fed into the decoder. 
% \citet{goodman-etal-2020-teaforn} extended teacher forcing from decoding $1$ step to $N$ steps and allowed using model's prediction as input prefix, but it still used the ground truth reference sequence as the oracle as opposed to using a dynamic oracle.

More recently, confidence-aware scheduled sampling uses the model's predictions as supervision where it's confident and the ground truth as supervision otherwise  \cite{liu-etal-2021-confidence}. Studies find that increasing the probability of choosing the model's prediction as supervision in later decoding steps outperforms vanilla scheduled sampling
\cite{liu-etal-2021-scheduled-sampling}.  Furthermore, we can randomly mask out some tokens during decoding and use an auxiliary task to recover the masked-out tokens, which yields performance gains \cite{ijcai2021p534}. Elastic weight consolidation can also be used to enhance the model's reliance on its previous outputs during decoding \cite{korakakis-vlachos-2022-improving}

\subsection{Imitation Learning}

SEARN casts learning as reinforcement learning, using the model's own prediction to generate examples and then searching over the action space to compute each action's cost \cite{Daum2009SearchbasedSP}. The LOLS and MIXER algorithms improve upon SEARN \cite{Chang2015LearningTS,RanzatoCAZ15}. MIXER additionally allowed mixed ground truth and model prediction to be used as the prefix, which is particularly relevant to this study.  

\subsection{Global-Aware Training}

An issue with autoregressive generation in seq2seq models is the underutilization of target-side future information during training. To address this, \citet{9989330} suggest learning the distribution of future translations at each decoding step. The global-aware beam search algorithm takes global attention distribution into account.
The seer-forcing algorithm employs an auxiliary Seer module, which captures future target information and guides the decoder during teacher forcing. \cite{feng-etal-2021-guiding}

\subsection{Alternative Loss Objectives}

Beyond MLE, there exist alternative loss functions to tackle exposure bias. For instance, minimum-risk training samples a batch of potential translations, and formulates the loss as as the sum of evaluation metrics (e.g., BLEU score) for each sample translation, weighted by the model-assigned probabilities \cite{shen-etal-2016-minimum}. Another study employs beam search during training and defines the loss as the frequency with which the gold sequence falls off the beam \cite{wiseman-rush-2016-sequence}. Furthermore, the mixed cross-entropy loss allows each token to have multiple sources of supervision, which can include the ground truth or the model's previous output \cite{li2021mixed}. To enhance robustness, we could perturb the input and regularize the model to minimize the output difference between the original and perturbed input \citet{guo-etal-2022-prediction}

\section{Discussion \& Future Work}

\paragraph{Pre-trained Models}
Our experiments for NER and summarization showed DAgger to be very effective at fine-tuning pre-trained models like BERT, FLAIR, and BART. However, it failed when we trained the Transformer from scratch for MT. As the original DAgger work suggests \cite{ross2011reduction}, the role of initial training of the model policy seems to determine the overall effectiveness of the method.

\paragraph{Runtime}
The beam search for dynamic completion can be optimized based on the specific metric. For example, for ROUGE and BLEU, we can cache the $n$-gram counts as we expand the beam search. We could also seek to improve searching for approximate dynamic oracles by drawing ideas from existing works \citep{dreyer-etal-2007-comparing, sokolov2012computing} that use other approximations in searching for optimal completion of BLEU.

\paragraph{Stochastic Dynamic Oracles}
In certain scenarios, it's possible to encounter multiple dynamic oracles with equivalent scores. In such cases, one approach could involve randomly selecting a completion from this set of equally-scored oracles to serve as the supervision for training. This data augmentation strategy has the potential to stimulate the model to produce more diverse outputs, fostering improved generalization.

\paragraph{Other Metrics and Models} While we focused on conventional metrics, our proposed algorithm could be extended to encompass more sophisticated metrics, including model-based ones like BERTScore \cite{zhang2020bertscore}. Furthermore, although we primarily employed encoder-decoder models in this study, our proposed algorithm can seamlessly be applied to decoder-only models without any modifications.

\section{Conclusion}

In this work, we identified deficiencies of commonly used training techniques for sequence training, including teacher forcing and scheduled sampling, and proposed to use DAgger training with loss-specific dynamic oracles. We designed novel dynamic oracle algorithms as a demonstration, exact dynamic oracles for decomposable metrics like exact-match F1 and partial-match F1, and approximate dynamic oracles based on beam search for non-decomposable metrics like ROUGE and BLEU. We empirically verified the effectiveness of exact and approximate dynamic oracles on three metrics and three tasks: partial-match F1 for named entity recognition, ROUGE for summarization, and BLEU for machine translation. Our results showed that seq2seq models trained by DAgger with exact and approximate dynamic oracles yield less favorable performance on machine translation, but outperform existing training techniques on named entity recognition and summarization. 

% can improve performance compared to existing training techniques.

\section{Limitations}

In practice, the runtime of the dynamic oracle for BLEU score is approximately 6 times longer than that of teacher forcing, mainly due to the beam search process. However, we can accelerate the beam search procedure through multi-threading, albeit at the expense of increased CPU memory usage. Considering that CPU memory costs are low, and the additional overhead is only incurred during the training phase, our methods remain practical for deployment.

In the case of non-decomposable metrics, the no-regret guarantee is forfeited because the dynamic oracle becomes approximate. This leaves space for future work to explore algorithms for approximating the dynamic oracle that approach optimality more closely. 
%For instance, instead of relying on beam search, one could explore training a model to predict the oracle word or employ methods rooted in dynamic programming (DP). We leave those to future work. 

% no theoratical guarantee for approaximate DO; but could potentially do a looser condition/empirically validate this. We leave this to future work. 

\bibliography{custom}
\bibliographystyle{acl_natbib}

\clearpage

\appendix

\section{Appendix}
\label{sec:appendix}

\subsection{Partial F1 Score Algorithm Proof of Correctness}
\label{app:partial_f1_proof}

% \harshine{proof of correctness done by proof with examples}

Algorithm \ref{algo:do_f1} represents our dynamic oracle for partial F1 Score. Here we give a proof of correctness to this algorithm.

We enumerate all possible cases, and prove each case for its correctness.

\begin{itemize}
    \item[] \textbf{\textit{Case 1:}} If the tagging has already begun, with the same type, \textit{i.e}, \\
    \texttt{prev\_gold\_tag = 'O' \\
            curr\_gold\_tag = 'B-LOC' \\ 
            prev\_tag = 'B-LOC'}\\
    to score partial match score, the best next tag should be \texttt{I-LOC}.
    \item[] \textbf{\textit{Case 2:}} If the tagging has not started or if there is type mismatch in the previous tag, \textit{i.e}, \\
    \texttt{prev\_gold\_tag = 'O' \\
            curr\_gold\_tag = 'B-LOC' \\ 
            prev\_tag = 'I-ORG'}\\
    to score partial or exact match score, the best next tag should be \texttt{B-ORG}.
    \item[] \textbf{\textit{Case 3:}} If the tagging has not already begun, but in the gold sequence, the tagging has already started, \textit{i.e}, \\
    \texttt{prev\_gold\_tag = 'B-LOC' \\
            curr\_gold\_tag = 'I-LOC' \\ 
            prev\_tag = 'O'}\\
    to score partial match score, the best next tag should be \texttt{B-LOC}. 
    \item[] \textbf{\textit{Case 4:}} If the previous tag type does not match the current tag type, \textit{i.e}, \\
    \texttt{prev\_gold\_tag = 'B-LOC' \\
            curr\_gold\_tag = 'I-LOC' \\ 
            prev\_tag = 'B-PER'}\\
    to not reduce the F1 Score due to False Negatives or False Positives, the best next prediction can either be \texttt{I-PER} or \texttt{O}, and the safest case would be to predict \texttt{O}.
\end{itemize}

\subsection{Exact Match F1 Score}
Early work in chunking (CoNLL-2000 shared task \cite{tjong_kim_sang_introduction_2000}) and NER (CoNLL-2003 shared task \cite{tjong_kim_sang_introduction_2003}) established F1 as the primary metric for evaluating span-based sequence tagging tasks. Here we refer to this metric as \emph{exact-match} F1 because it computes the harmonic mean of precision and recall based on only spans which exactly match (both in the span of tokens and the label) between the reference and system output.

Algorithm \ref{algo:do_f1_exact} presents our dynamic oracle for exact-match F1.
This is a very simple algorithm for predicting the next best token for the NER task.

\begin{algorithm}[t]
  \DontPrintSemicolon
  \SetNoFillComment
  \SetKwInOut{Input}{Input}
  \SetKwInOut{Output}{Output}
  \Input{$\mathbf{prev\_gold\_tag}$: previous gold tag,\\
   $\mathbf{curr\_gold\_tag}$: current gold tag, \\
   $\mathbf{prev\_tag}$: previous predicted single tag}
  \Output{next best }

    \uIf{start(\textup{curr\_gold\_tag}) = `B'}{
        \uIf{\textup{prev\_tag} $\neq$ `O'}{
            \Return `O'
        }
        \Else{
            \Return $\mathbf{curr\_gold\_tag}$
        }
    }
    \uElseIf{start(\textup{curr\_gold\_tag}) = `I'}{
        \uIf{\textup{prev\_tag} is `O'}{
            \Return `O'
        }
        \uElseIf{type(\textup{prev\_tag}) $\neq$ type(\textup{prev\_gold\_tag}) and start(\textup{prev\_tag}) $\neq$ start(\textup{prev\_gold\_tag}) }{
            \Return `O'
        }
        \Else{
            \Return $\mathbf{curr\_gold\_tag}$
        }
    }
    \Else{
        \Return `O'
    }
    \textbf{Note:} 
    \begin{itemize}
    \item \textit{start}(tag) returns the prefix of the tag, \\
    i.e, \texttt{B}, \texttt{I}, or \texttt{O}.
    \item \textit{type}(tag) returns the entity type, i.e, \texttt{PER}, \texttt{LOC}, etc.
    \end{itemize}
  
 \caption{Dynamic Oracle for exact F1 Score}
 \label{algo:do_f1_exact}
\end{algorithm}

\subsection{Exact F1 Score algorithm}

The dynamic oracle algorithm for exact F1 score algorithm is given in Algorithm \ref{algo:do_f1_exact}.

\section{Word error rate (WER)}
\label{app:wer}

During our study, we have also explored dynamic oracle for Word Error Rate (WER), which is commonly used metric in Automatic Speech Recognition (ASR) tasks. We decided not to put this section in the main text because we later found there has been a concurrent and similar research \citep{sabour2018optimal}. Nonetheless, in hope of helping future researchers, here we present our dynamic oracle for WER (Section \ref{app:wer_oracle}) and a proof of correctness (Section \ref{app:wer_proof}).

\subsection{Dynamic oracle}
\label{app:wer_oracle}

% TODO: figure out how to have a compact version in the main section

% \begin{figure*}
% \removelatexerror
% \centering

\begin{algorithm*}
  \DontPrintSemicolon
  \SetNoFillComment
  \SetKwInOut{Input}{Input}
  \SetKwInOut{Output}{Output}
  \Input{$\mathbf{G}$: token array[$L$] - ground truth sequence,\\
   $\mathbf{S}$: token array[$L$] - previous tokens fed to model}
  \Output{\textbf{O}: token array[$L$] - dynamic oracle sequence}

  \SetKwFunction{FMain}{DynamicOracleWER}
  \SetKwProg{Fn}{Function}{:}{}
  \Fn{\FMain{$\mathbf{G}$,$\mathbf{S}$}}{
    $\mathbf{DP_{WER}}\gets$ empty int array[$L+1$][$L+1$]\;
    $\mathbf{DP_{D.O.}}\gets$ int array[$L+1$][$L+1$] with $+\infty$'s\;
    set $\mathbf{DP_{WER}}_{[0][:]}, \mathbf{DP_{D.O.}}_{[0][:]}$ to be 0\dots $L$   \tcp*{for the start token}
    set $\mathbf{DP_{WER}}_{[:][0]}, \mathbf{DP_{D.O.}}_{[:][0]}$ to be 0\dots $L$  \tcp*{for the start token}\;

    \For(\tcp*[f]{loop over rows}){$i\gets1$ \KwTo $L+1$}{
      \For(\tcp*[f]{loop over columns}){$j\gets1$ \KwTo $L+1$}{
        \tcc{calculating WER for each candidate token, using dynamic programming}
        \uIf{$\mathbf{S}_{[i]}$ = $\mathbf{G}_{[j]}$}{
          $\var{penalty} \gets 0$
        }
        \Else{
          $\var{penalty} \gets 1$
        }
        $\mathbf{DP_{WER}}_{[i][j]} \gets \min \Bigg\{\substack{
        \mathbf{DP_{WER}}_{[i-1][j-1]} + \var{penalty},\\
        \mathbf{DP_{WER}}_{[i-1][j]} + 1,\\
        \mathbf{DP_{WER}}_{[i][j-1]} + 1}\Bigg\} $  \;
        
        $\mathbf{DP_{D.O.}}_{[i][j]} \gets \min \Bigg\{\substack{
        \mathbf{DP_{WER}}_{[i-1][j-1]},\\
        \mathbf{DP_{WER}}_{[i-1][j]} + 1,\\
        \mathbf{DP_{D.O.}}_{[i][j-1]} + 1}\Bigg\} $  \;
      }
      \var{idx} $\gets \argmin\{\mathbf{DP_{D.O.}}_{[i][:]}\}$ \tcp*{select token with the lowest WER as label}
      \uIf{\textup{\var{idx}} = $L+1$}{
        $\mathbf{O}_{[i]} \gets$ $\langle \text{End}\rangle$ \tcp*{further growing the chart will not reduce WER score, so output $\langle \text{End}\rangle$}
      }
      \Else{
        $\mathbf{O}_{[i]} \gets \mathbf{G}_{\text{\var{idx}}}$ \;
      }
    }
    \Return $\mathbf{O}_{[1:L+1]}$\;
  }
  \caption{Dynamic Oracle for Word Error Rate}
 \label{algo:do_wer}
\end{algorithm*}
% \caption{Dynamic Oracle for Word Error Rate}
% \label{fig: do_wer}
% \end{figure*}

\paragraph{Note on Tie-breaking} Note that on line 15, there exist multiple ways to break ties. Currently we use the following tie breaking preference:
If the ground truth index is among the minima, select the ground truth index; if not, then select the largest index (this effectively tells the dynamic oracle sequence to prefer shorter sequence).

\paragraph{Note on $S$} It is important to understand the meaning of $S$ here. $S$ is the tokens fed to the model. If we are using Teacher Forcing (or equivalently, Scheduled Sampling with sampling probability of zero), then $S$ is just $G$; if we are using Scheduled Sampling with non-zero sampling probability, then $S$ contains both ground truth tokens and model predicted tokens; if we are using Scheduled Sampling with sampling probability of 1, then $S$ is just the model's predicted sequence. 

\paragraph{Note on modularization and optimization} Note that the above algorithm can be executed in a post-hoc fashion, meaning that we can calculate the Dynamic Oracle after the entire GPU forwarding is finished. This modularization allows us to port the entire Dynamic Oracle implementation to C as opposed to native Python, which makes things faster; also, this formulation modularizes our method as a plug-and-play component for sequence generation tasks and allows it to serve as an easy extension to existing systems.\\

\subsection{Proof of Correctness}
\label{app:wer_proof}

\textbf{Line 14:} \\
\begin{enumerate}
    \item \textit{Choosing A is always superior to choosing $\sim$A, in that choosing A will lead to no penalty (i.e) substitution error.} \\ \\ 
    If the $j^{th}$ word in the gold sequence matches with the $i^{th}$ word in the predicted sequence, there is no substitution that has to be done. But if the words do not match, the $j^{th}$ word in the gold sequence will have to be replaced with the $i^{th}$ word in the predicted sequence. \\ \\
    If the $j^{th}$ word in the gold sequence matches with the $i^{th}$ word in the predicted sequence, the substitution WER does not increase. \\ \\
    If the $j^{th}$ word in the gold sequence does not match with the $i^{th}$ word in the predicted sequence, the substitution WER will increase by a value of 1 when compared to the WER of the sequence up to that point, (i.e), the WER until the $(j-1)^{th}$ word in the gold sequence corresponding to the $(i-1)^{th}$ word in the predicted sequence. \\ 
    A penalty of 1 is added to the WER if the words do not match. So, here to get rid of this penalty term when calculating the substitution error, we always select the same word as in the gold sequence when we are trying to extend the sequence. \\ \\
    For example, if we consider the gold sequence as \texttt{A B C} and the current predicted sequence is \texttt{A B}. The chart is shown in Table \ref{tab:example_chart}. 
    
    \begin{table}[t]
        \centering
        \begin{tabular}{c|c c c c}
             & \textbf{$\epsilon$} & \texttt{A} & \texttt{B} & \texttt{C} \\
            \hline
            \textbf{$\epsilon$} & 0 & 1 & 2 & 3 \\
            \texttt{A} & 1 & 0 & 1 & 2 \\
            \texttt{B} & 2 & 1 & 0 & 1 \\
        \end{tabular}
        \caption{Chart for Gold Sequence \texttt{A B C} and Predicted Sequence \texttt{A B}}
        \label{tab:example_chart}
    \end{table}
    
    By choosing \texttt{A}, \texttt{B} and \texttt{C} for each cell respectively when growing the chart, the substitution errors will be, \\
    
        \texttt{Substitution Error for A = 2} \\
        \texttt{Substitution Error for B = 1} \\ 
        \texttt{Substitution Error for C = 0} \\
    
    By choosing $\sim$\texttt{A}, $\sim$\texttt{B} and $\sim$\texttt{C} for each cell respectively when growing the chart, the substitution errors will be, \\ 
    
        \texttt{Substitution Error for $\sim$A = 3} \\
        \texttt{Substitution Error for $\sim$B = 2} \\ 
        \texttt{Substitution Error for $\sim$C = 1} \\ 
    
    Therefore, the reason for we prefer choosing A over $\sim$A here is because A (choosing the same word as the gold sequence) does not incur any substitution error, whereas $\sim$A (choosing a different word) always incurs a substitution error. So, choosing the same word is always superior. \\
    
    \item\textit{The terms $DP_{W.E.R[i-1][j]} + 1$ and $DP_{D.O[i][j-1]} + 1$ correspond to insertion and deletion errors respectively.} \\ \\
    If the $i^{th}$ word has to be inserted to the sequence, then the WER will increase by 1 when compared to the WER corresponding to the sequence until that point (i.e) the WER until the $j^{th}$ word in the gold sequence corresponding to the $(i-1)^{th}$ word in the predicted sequence. \\ \\
    If the $i^{th}$ word has to be deleted from the sequence, then the WER will increase by 1 when compared to the WER corresponding to the sequence until that point (i.e) the WER until the $(j-1)^{th}$ word in the gold sequence corresponding to the $i^{th}$ word in the predicted sequence. \\ \\ \\
    For example, if the gold sequence is \texttt{A B C} and the predicted sequence until now is \texttt{A B}, \\
    
        \texttt{WER to insert C = WER corresponding to AB + 1} \\
    
    If the gold sequence is \texttt{A B} and the predicted sequence until now is \texttt{A B C}, \\
    
        \texttt{WER to delete C = WER corresponding to ABC + 1} \\

    Therefore we can say that the second term corresponds to the insertion error term and the third term corresponds to the deletion error term when calculating the WER. \\
\end{enumerate}
Since these 3 operations are the only possible operations that we could do, and for each of the 3 we have calculated the overall lowest cost of doing that thing, the overall best thing to do must be the best of the 3 of them (i.e) the minimum of these values. \\ \\

\textbf{Line 15:} \\
\textit{Choosing $G_{ind}$, where $ind = argmin(DP_{D.O [i]})$, as the first token in the gold sequence is superior to choosing $G_{ind'}$, where $ind \neq ind'$, as the start token for the gold sequence in that $G_{ind}$ leads to a sequence of least possible WER.}

\begin{enumerate}
    \item \textbf{Case 1:} There are no ties (i.e) there is only one optimal solution in that particular step. \\ \\
    Let us assume instead of choosing the argmin WER, we choose some other word (i.e) we choose some other word at index $ind'$ and $ind' \neq ind$ where $ind = argmin(DP_{O [i]})$. \\ \\
    The $DP_{W.E.R}$ is adjusted in a way that the word that has been chosen is the word corresponding to the word at index $ ind'$. \\ \\ 
    We know that at every step, there is a possibility of the best WER at the previous step to increase by 1 or not increase at all. \\
    The best WER will remain the same if the next predicted word matches the word in the gold sequence. In this case, there is no penalty added. \\
    If the words do not match, a penalty will have to be added (either as an insertion error or as a substitution error), as explained in the previous proof.
    \begin{enumerate}
        \item Choosing the non-optimal word \\ 
        Let us denote the WER at the $i^{th}$ predicted word for the index \textbf{ind'} as b. \\ \\
        Upon further growing the chart, \\
        b $\leq$ minimum WER at the $(i+1)^{th}$ predicted word $\leq$ (b + 1)\\
        . \\
        . \\
        . \\
        b $\leq$ minimum WER at the $(i+n)^{th}$ predicted word $\leq$ (b + n)\\
        where n denoted the number of steps. \\ \\
        If we keep choosing the optimal word at all the following steps and stop the sequence once the stopping condition is reached, the WER will not increase. \\
        If we always choose something other than the optimal word at all the following steps, the best score we can get will be $b + n$, assuming there are n steps.
        \item Choosing the optimal word (i.e) the word corresponding to the argmin of the WER \\
        Let us denote the WER at the $i^{th}$ predicted word for the index \textbf{ind} as a, where a $<$ b, since a corresponds to the argmin word. \\ \\
        Upon further growing the chart, \\
        a $\leq$ minimum WER at the $(i+1)^{th}$ predicted word $\leq$ (a + 1)\\
        . \\
        . \\
        . \\
        a $\leq$ minimum WER at the $(i+n)^{th}$ predicted word $\leq$ (a + n)\\
        where n denoted the number of steps. \\ \\
        Similar to the previous selection, if we keep choosing the optimal word at all the following steps and stop the sequence once the stopping condition is reached, the WER will not increase. \\
        If we always choose something other than the optimal word at all the following steps, the best score we can get will be $a + n$, assuming there are n steps.
    \end{enumerate}
    Assuming that we have only made optimal word selection until the $i^{th}$ word, and we will only be making optimal selection after the $i^{th}$ word, the final WER will be \texttt{a} if we choose the word corresponding to the index \textbf{ind} (the argmin WER at the $i^{th}$ step) and the WER will be \texttt{b} if we choose the word corresponding to the index \textbf{ind'}(a word which is different from the argmin WER at the $i^{th}$ step), which is not the optimal path as another path with lower WER exists.
    \item \textbf{Case 2:} There are ties (i.e) more than one optimal solution exists at a particular step. \\ \\
    Let us assume we have argmin values at indices $j$, $j+m$ and $j+n$, where m, n $\in$ \textbf{Z}. \\ \\
    With a proof similar to the previous case, having chosen the optimal word upto the the $i^{th}$ step and we can say that by choosing any of the optimal words at the $i^{th}$ step and by continuously choosing the optimal word at the following steps, we will end up with the same WER. \\
    By choosing a word at any of these minimum indices, we will reach the same optimal WER at the end. \\ \\
    In this implementation, we are choosing the word corresponding to the minimum index as it would help in matching the sequence lengths the best. 
\end{enumerate}
\textbf{Stopping Condition:} \\
\textit{If the minimum WER corresponds to the last word in the gold sequence, the end of the sequence has been reached.} \\ \\
When the minimum WER corresponds to the last word in the gold sequence (i.e) $min\_WER = DP_{D.O [i][L + 1]}$, extending the sequence further will only increase the WER. \\ \\ 
Let us assume we are going to grow the chart further after reaching the minimum WER at the last word, the argmin will once be the last word. We know that,
$DP_{D.O [i+1][j]} = min(DP_{W.E.R [i][j-1]} , DP_{W.E.R [i][j]} + 1 , DP_{D.O [i+1][j-1]} + 1)$ \\ \\
In the $i^{th}$ row, $DP_{W.E.R [i][j]} > DP_{W.E.R [i][L+1]}$ for all $0 \leq j < L+1$. \\
From the previous proofs, we can say that in the $(i+1)^{th}$ row, $DP_{D.O [i+1][j]} > DP_{W.E.R [i][L+1]}$ for all $0 \leq j \leq L+1$. \\ \\
This causes an increase in the optimal WER for the final sequence. \\
Therefore the condition for stopping when the minimum WER corresponds to the last word in the gold sequence is correct.

% \subsection{Other Results on NER}

% \input{tables/tex/ner_appendix}

\end{document}